\def\year{2019}\relax
\documentclass[letterpaper]{article} 
\usepackage{aaai19}  
\usepackage{times}  
\usepackage{helvet}  
\usepackage{courier}  
\usepackage{url}  
\usepackage{graphicx}  

\usepackage{algorithm}
\usepackage{algpseudocode}
\usepackage{amssymb}
\usepackage{bbding}
\usepackage{amsmath}
\usepackage{tabularx, booktabs} 

\newcommand{\LL}{\mathcal{L}}
\newcommand{\EE}{\mathop{\mathbb{E}}}

\begin{document}
%
\title{MEAL: Multi-Model Ensemble via Adversarial Learning}
\author{Zhiqiang Shen$^{1,2*}$,~~Zhankui He$^{3}$\thanks{Equal contribution. This work was done when Zhankui He was a research intern at University of Illinois at Urbana-Champaign.},~~Xiangyang Xue$^{1}$\\
	$^1$Shanghai Key Laboratory of Intelligent Information Processing, \\ School of Computer Science, Fudan University, Shanghai, China \\
	$^2$Beckman Institute, University of Illinois at Urbana-Champaign, IL, USA\\
	$^3$School of Data Science, Fudan University, Shanghai, China\\
	\tt\small zhiqiangshen0214@gmail.com,
	\tt\small \{zkhe15, xyxue\}@fudan.edu.cn
}
\maketitle
\begin{abstract}
Often the best performing deep neural models are ensembles of multiple base-level networks. Unfortunately, the space required to store these many networks, and the time required to execute them at test-time, prohibits their use in applications where test sets are large (e.g., ImageNet).
In this paper, we present a method for compressing large, complex trained ensembles into a single network,
where knowledge from a variety of trained deep neural networks (DNNs) is distilled and transferred to a single DNN. In order to distill diverse knowledge from different trained (teacher) models, we propose to use adversarial-based learning strategy where we define a block-wise training loss to guide and optimize the predefined student network to recover the knowledge in teacher models, and to promote the discriminator network to distinguish teacher {\em vs.} student features simultaneously. The proposed ensemble method (MEAL) of transferring distilled knowledge with adversarial learning exhibits three important advantages: (1) the student network that learns the distilled knowledge with discriminators is optimized better than the original model; (2) fast inference is realized by a single forward pass, while the performance is even better than traditional ensembles from multi-original models; (3) the student network can learn the distilled knowledge from a teacher model that has arbitrary structures. 
Extensive experiments on CIFAR-10/100, SVHN and ImageNet datasets demonstrate the effectiveness of our MEAL method. On ImageNet, our ResNet-50 based MEAL achieves top-1/5 21.79\%/5.99\% val error, which outperforms the original model by 2.06\%/1.14\%. Code and models are available at: \url{https://github.com/AaronHeee/MEAL}. 
\end{abstract}

\section{1.~~Introduction}

The ensemble approach is a collection of neural networks whose predictions are combined at test stage by weighted averaging or voting.
It has been long observed that ensembles of multiple networks are generally much more robust and accurate than a single network. This benefit has also been exploited indirectly when training a single network through Dropout~\cite{srivastava2014dropout}, Dropconnect~\cite{wan2013regularization}, Stochastic Depth~\cite{huang2016deep}, Swapout~\cite{singh2016swapout}, etc. We extend this idea by forming ensemble predictions during training, using the outputs of different network architectures with different or identical augmented input. Our testing still operates on a single network, but the supervision labels made on different pre-trained networks correspond to an ensemble prediction of a group of individual reference networks.

\begin{figure}[t]
	\centering
	\includegraphics[width=0.43\textwidth]{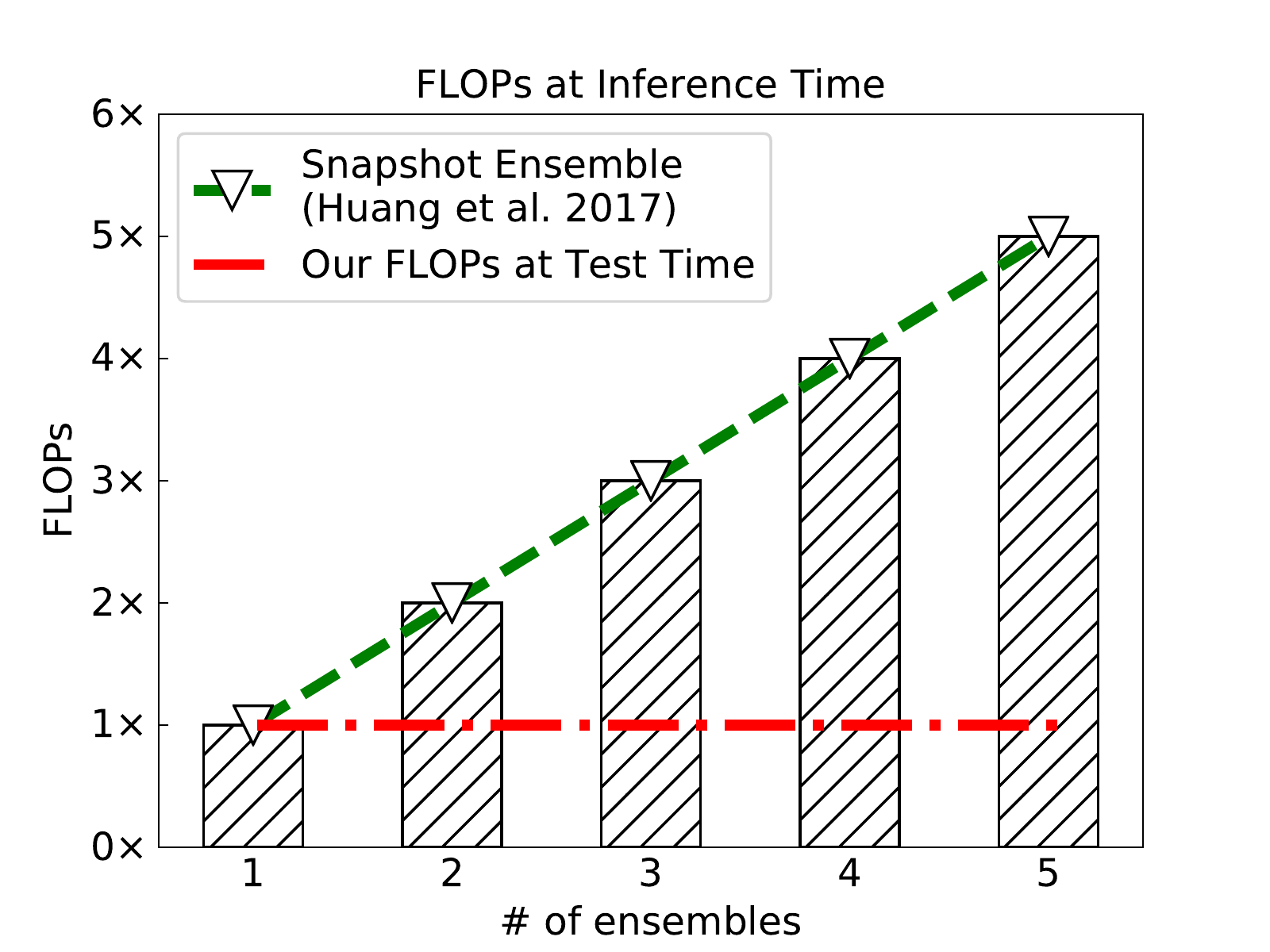}
	\vspace{-0.10in}
	\caption{Comparison of FLOPs at inference time. Huang et al.~\cite{huang2017snapshot} employ models at different local minimum for ensembling, which enables no additional training cost, but the computational FLOPs at test time linearly increase with more ensembles. In contrast, our method use only one model during inference time throughout, so the testing cost is independent of \# ensembles.}
	\label{d}
	\vspace{-0.1in}
\end{figure}

The traditional ensemble, or called true ensemble, has some disadvantages that are often overlooked. 1) Redundancy: The information or knowledge contained in the trained neural networks are always redundant and has overlaps between with each other. Directly combining the predictions often requires extra computational cost but the gain is limited. 2) Ensemble is always large and slow: Ensemble requires more computing operations than an individual network, which makes it unusable for applications with limited memory, storage space, or computational power such as  desktop, mobile and even embedded devices, and for applications in which real-time predictions are needed.

\begin{figure}[t]
	\centering
	\includegraphics[width=0.48\textwidth]{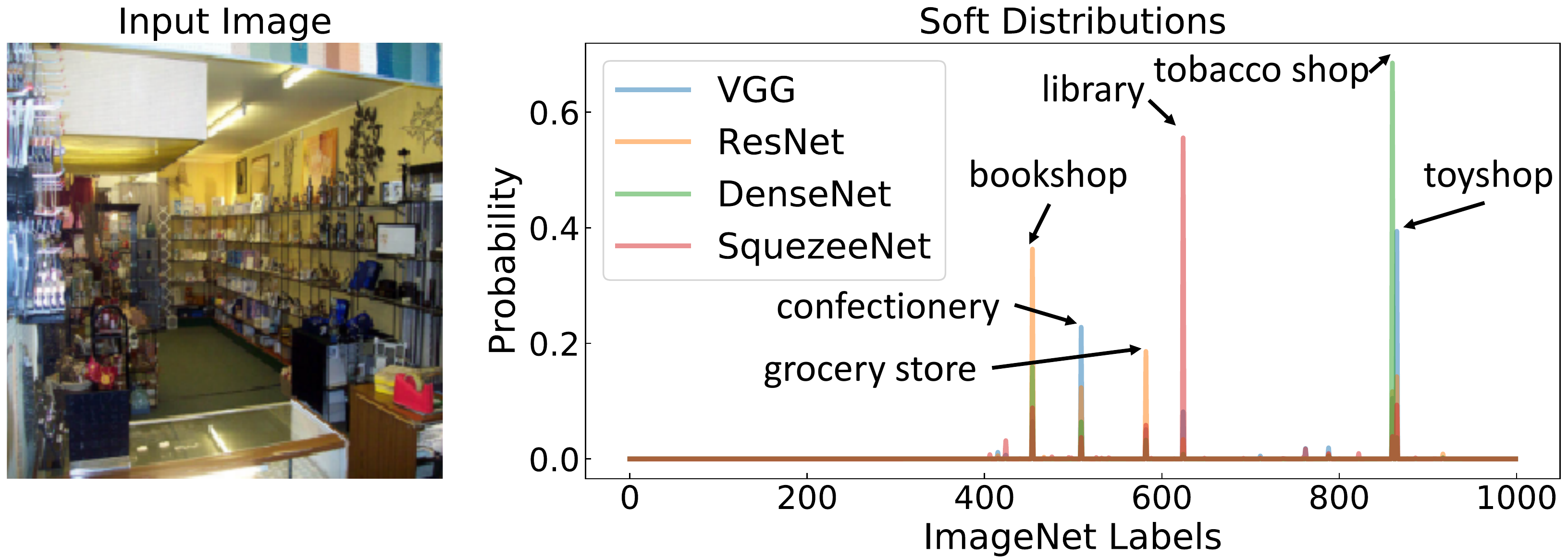}
	\vspace{-0.20in}
	\caption{Left is a training example of class ``tobacco shop'' from ImageNet. Right are soft distributions from different trained architectures. The soft labels are more {\em {informative}} and can provide more coverage for visually-related scenes.}
	\label{soft_labels}
	\vspace{-0.1in}
\end{figure}

To address the aforementioned shortcomings, in this paper we propose to use a learning-based ensemble method. Our goal is to learn an ensemble of multiple neural networks without incurring any additional {\em {testing costs}}. We achieve this goal by leveraging the combination of diverse outputs from different neural networks as supervisions to guide the target network training. The reference networks are called {\em {Teachers}} and the target networks are called {\em {Students}}. Instead of using the traditional one-hot vector labels, we use the {\em {soft}} labels that provide more coverage for co-occurring and visually related objects and scenes. We argue that labels should be informative for the specific image. In other words, the labels should not be identical for all the given images with the same class. More specifically, as shown in Fig.~\ref{soft_labels}, an image of ``tobacco shop'' has similar appearance to ``library'' should have a different label distribution than an image of ``tobacco shop'' but is more similar to ``grocery store''. It can also be observed that soft labels can provide the additional intra- and inter-category relations of datasets.

To further improve the robustness of student networks, we introduce an adversarial learning strategy to force the student to generate similar outputs as teachers.
Our experiments show that MEAL consistently improves the accuracy across a variety of popular network architectures on different datasets. For instance, our shake-shake~\cite{gastaldi2017shake} based MEAL achieves 2.54\% test error on CIFAR-10, which is a relative $11.2\%$ improvement\footnote{Shake-shake baseline~\cite{gastaldi2017shake} is 2.86\%.}. On ImageNet, our ResNet-50 based MEAL achieves 21.79\%/5.99\% val error, which outperforms the baseline by a large margin.

In summary, our contribution in this paper is three fold.

\begin{itemize}
	\addtolength{\itemsep}{-0.05in}
	\item  An end-to-end framework with adversarial learning is designed based on the {\em {teacher-student}} learning paradigm for deep neural network ensembling.
	\item The proposed method can achieve the goal of ensembling multiple neural networks with no additional {\em {testing cost}}.
	\item The proposed method improves the state-of-the-art accuracy on CIFAR-10/100, SVHN, ImageNet for a variety of existing network architectures.
\end{itemize}

\section{2.~~Related Work}

There is a large body of previous work~\cite{hansen1990neural,perrone1995networks,krogh1995neural,dietterich2000ensemble,huang2017snapshot,lakshminarayanan2017simple} on ensembles with neural networks. However, most of these prior studies focus on improving the generalization of an individual network. Recently, Snapshot Ensembles~\cite{huang2017snapshot} is proposed to address the cost of training ensembles. In contrast to the Snapshot Ensembles, here we focus on the cost of {\em {testing ensembles}}. Our method is based on the recently raised knowledge distillation~\cite{hinton2015distilling,papernot2016semi,li2017learning,yim2017gift} and adversarial learning~\cite{goodfellow2014generative}, so we will review the ones that are most directly connected to our work.

\noindent{\textbf{``Implicit'' Ensembling.}}
Essentially, our method is an ``implicit'' ensemble which usually has high efficiency during both training and testing. The typical ``implicit'' ensemble methods include: Dropout~\cite{srivastava2014dropout}, DropConnection~\cite{wan2013regularization}, Stochastic Depth~\cite{huang2016deep}, Swapout~\cite{singh2016swapout}, etc. These methods generally create an exponential number of networks with shared weights during training and then implicitly ensemble them at test time. In contrast, our method focuses on the  subtle differences of labels with identical input. Perhaps the most similar to our work is the recent proposed Label Refinery~\cite{bagherinezhad2018label}, who focus on the single model refinement using the softened labels from the previous trained neural networks and iteratively learn a new and more accurate network. Our method differs from it in that we introduce adversarial modules to force the model to learn the difference between teachers and students, which can improve model generalization and can be used in conjunction with any other implicit ensembling techniques. 

\noindent{\textbf{Adversarial Learning.}}
Generative Adversarial Learning~\cite{goodfellow2014generative} is proposed to generate realistic-looking images from random noise using neural networks. It consists of two components. One serves as a generator and another one as a discriminator. The generator is used to synthesize images to fool the discriminator, meanwhile, the discriminator tries to distinguish real and fake images. Generally, the generator and discriminator are trained simultaneously through competing with each other. In this work, we employ generators to synthesize student features and use discriminator to discriminate between teacher and student outputs for the same input image. An advantage of adversarial learning is that the generator tries to produce similar features as a teacher that the discriminator cannot differentiate. This procedure improves the robustness of training for student network and has applied to many fields such as image generation~\cite{johnson2018image}, detection~\cite{bai2018finding}, etc.

\begin{figure*}[t]
	\centering
	\includegraphics[width=0.75\textwidth]{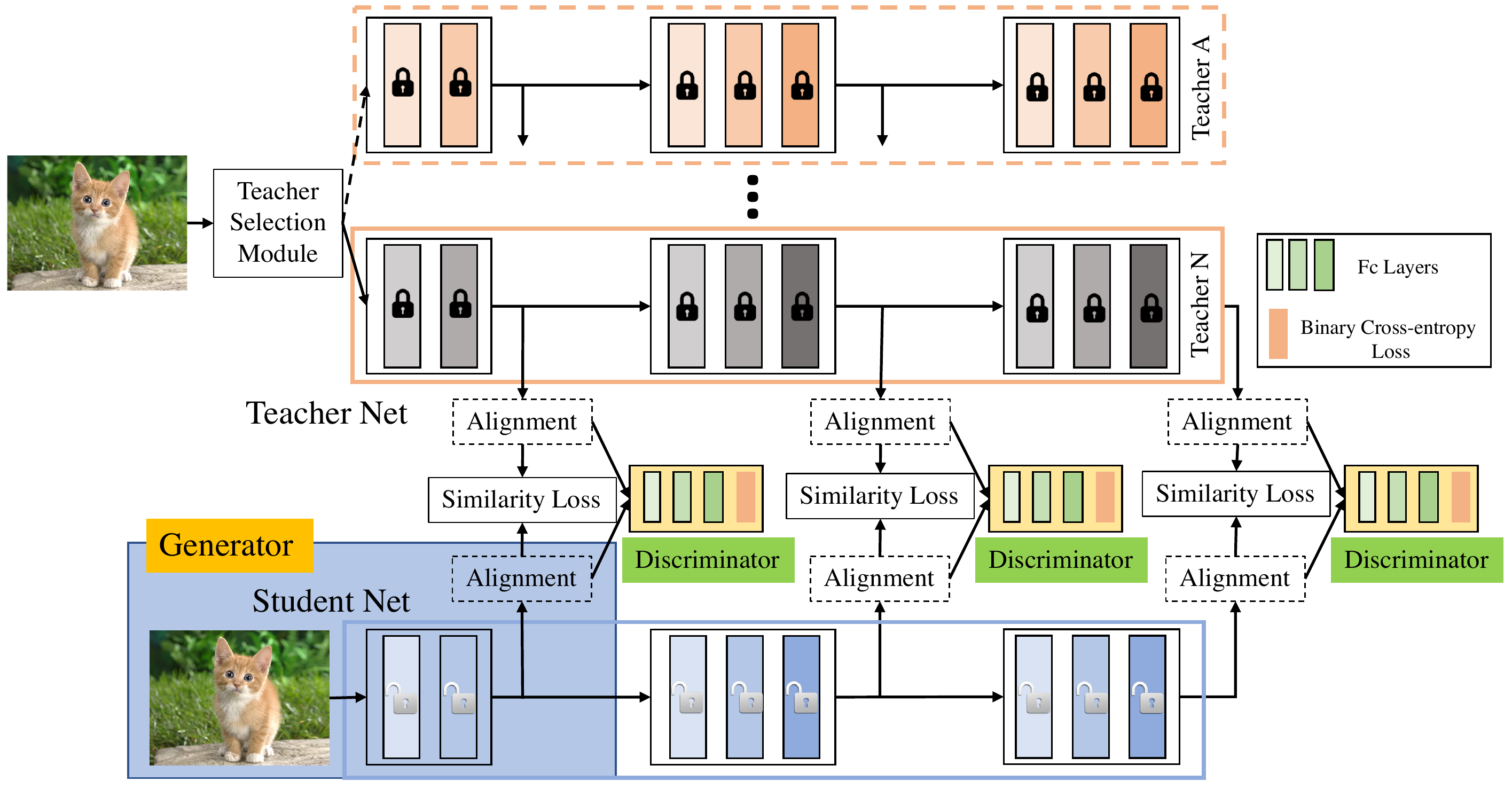}
	\vspace{-0.02in}
	\caption{Overview of our proposed architecture. We input the same image into the teacher and student networks to generate intermediate and final outputs for {\em {Similarity Loss}} and {\em {Discriminators}}. The model is trained adversarially against several {\em {discriminator networks}}. During training the model observes supervisions from trained teacher networks instead of the one-hot ground-truth labels, and the teacher's parameters are fixed all the time.}
	\label{al}
	\vspace{-0.08in}
\end{figure*}

\noindent{\textbf{Knowledge Transfer.}}
Distilling knowledge from trained neural networks and transferring it to another new network has been well explored in~\cite{hinton2015distilling,chen2015net2net,li2017learning,yim2017gift,bagherinezhad2018label,anil2018large}. The typical way of transferring knowledge is the {\em {teacher-student learning paradigm}}, which uses a softened distribution of the final output of a teacher network to teach information to a student network. With this teaching procedure, the student can learn how a teacher studied given tasks in a more efficient form. Yim et al.~\cite{yim2017gift} define the distilled knowledge to be transferred flows between different intermediate layers and computer the inner product between parameters from two networks. Bagherinezhad et al.~\cite{bagherinezhad2018label} studied the effects of various properties of labels and introduce the {\em {Label Refinery}} method that iteratively updated the ground truth labels after examining the entire dataset with the {{teacher-student learning paradigm}}.

\section{3.~~Overview}
\noindent{\textbf{Siamese-like Network Structure}}
Our framework is a siamese-like architecture that contains two-stream networks in teacher and student branches. The structures of two streams can be identical or different, but should have the same number of blocks, in order to utilize the intermediate outputs. 
The whole framework of our method is shown in Fig.~\ref{al}. It consists of a teacher network, a student network, alignment layers, similarity loss layers and discriminators.

The teacher and student networks are processed to generate intermediate outputs for alignment. The alignment layer is an adaptive pooling process that takes the same or different length feature vectors as input and output fixed-length new features. We force the model to output similar features of student and teacher by training student network adversarially against several discriminators.
We will elaborate each of these components in the following sections with more details.
\section{4.~~Adversarial Learning (AL) for Knowledge Distillation}

\subsection{4.1~~Similarity Measurement}
Given a dataset $\mathcal{D}={(X_i, Y_i)}$, we pre-trained the teacher network $\mathcal{T}_\theta$ over the dataset using the cross-entropy loss against the one-hot image-level labels{\footnote{Ground-truth labels}} in advance. The student network $\mathcal{S}_\theta$ is trained over the same set of images, but uses labels generated by $\mathcal{T}_\theta$. More formally, we can view this procedure as training $\mathcal{S}_\theta$ on a new labeled dataset $\tilde{\mathcal{D}}={(X_i, \mathcal{T}_\theta(X_i))}$. Once the teacher network is trained, we freeze its parameters when training the student network. 

We train the student network $\mathcal{S}_\theta$ by minimizing the similarity distance between its output and the soft label generated by the teacher network. Letting $p_c^{{\mathcal T}_\theta}({X_i})={\mathcal{T}_\theta }({X_i})[c]$, $p_c^{{\mathcal S}_\theta}({X_i})={\mathcal{S}_\theta }({X_i})[c]$ be the probabilities assigned to class $c$ in the teacher model $\mathcal T_\theta$ and student model $\mathcal S_\theta$. The similarity metric can be formulated as:

\begin{equation}
\begin{gathered}
\LL_{Sim} = d({\mathcal{T}_\theta }({X_i}),{\mathcal{S}_\theta }({X_i})) \hfill \\
\quad \quad \;\; = \sum\limits_c {d(p_c^{\mathcal{T}_\theta}({X_i}),p_c^{\mathcal{S}_\theta}({X_i}))}  \hfill \\ 
\end{gathered} 
\end{equation}

We investigated three distance metrics in this work, including $\ell_1$, $\ell_2$ and KL-divergence. The detailed experimental comparisons are shown in Tab.~\ref{ablation}. Here we formulate them as follows.

\noindent{\textbf{$\ell_1$ distance}} is used to minimize the absolute differences between the estimated student probability values and the reference teacher probability values. Here we formulate it as:
\begin{equation}
\LL_{\ell_1\_Sim}({\mathcal{S}_\theta }) = \frac{1}{n}\sum\limits_c {\sum\limits_{i = 1}^n {{{\left| {p_c^{\mathcal{T}_\theta }({X_i}) - p_c^{\mathcal{S}_\theta }({X_i})} \right|}^1}} } 
\end{equation}

\noindent{\textbf{$\ell_2$ distance}} or euclidean distance is the straight-line distance in euclidean space. We use $\ell_2$ loss function to minimize the error which is the sum of all squared differences between the student output probabilities and the teacher probabilities. The $\ell_2$ can be formulated as:
\begin{equation}
\LL_{\ell_2\_Sim}({\mathcal{S}_\theta }) = \frac{1}{n}\sum\limits_c {\sum\limits_{i = 1}^n {{{\left\| {p_c^{\mathcal{T}_\theta }({X_i}) - p_c^{\mathcal{S}_\theta }({X_i})} \right\|}^2}}} 
\end{equation}

\noindent{\textbf{KL-divergence}} is a measure of how one probability distribution is different from another reference probability distribution. Here we train student network $\mathcal{S}_\theta$ by minimizing the KL-divergence between its output $p_c^{{\mathcal S}_\theta}({X_i})$ and the soft labels $p_c^{{\mathcal T}_\theta}({X_i})$ generated by the teacher network. Our loss function is:
\begin{equation}
\begin{gathered}
{\LL_{KL\_Sim}}({\mathcal{S}_\theta }) =  - \frac{1}{n}\sum\limits_c {\sum\limits_{i = 1}^n {p_c^{{\mathcal{T}_\theta }}({X_i})\log (\frac{{p_c^{{S_\theta }}({X_i})}}{{p_c^{{\mathcal{T}_\theta }}({X_i})}})} }  \hfill \\
\quad \quad \quad  \quad \quad \quad    =  - \frac{1}{n}\sum\limits_c {\sum\limits_{i = 1}^n {p_c^{{\mathcal{T}_\theta }}({X_i})\log } } p_c^{{\mathcal{S}_\theta }}({X_i}) \hfill \\
\quad \quad \quad  \quad \quad \quad \quad  + \frac{1}{n}\sum\limits_c {\sum\limits_{i = 1}^n {p_c^{{\mathcal{T}_\theta }}({X_i})\log } } p_c^{{\mathcal{T}_\theta }}({X_i}) \hfill \\ 
\end{gathered} 
\end{equation}

where the second term is the entropy of soft labels from teacher network and is constant with respect to $\mathcal{T}_\theta$. We can remove it and simply minimize the cross-entropy loss as follows:

\begin{equation}
{{{\LL}}_{CE\_Sim}}({\mathcal{S}_\theta }) =  - \frac{1}{n}\sum\limits_c {\sum\limits_{i = 1}^n {p_c^{{\mathcal{T}_\theta }}({X_i})\log } } p_c^{{\mathcal{S}_\theta }}({X_i})
\end{equation}

\begin{figure}[t]
	\centering
	\includegraphics[width=0.48\textwidth]{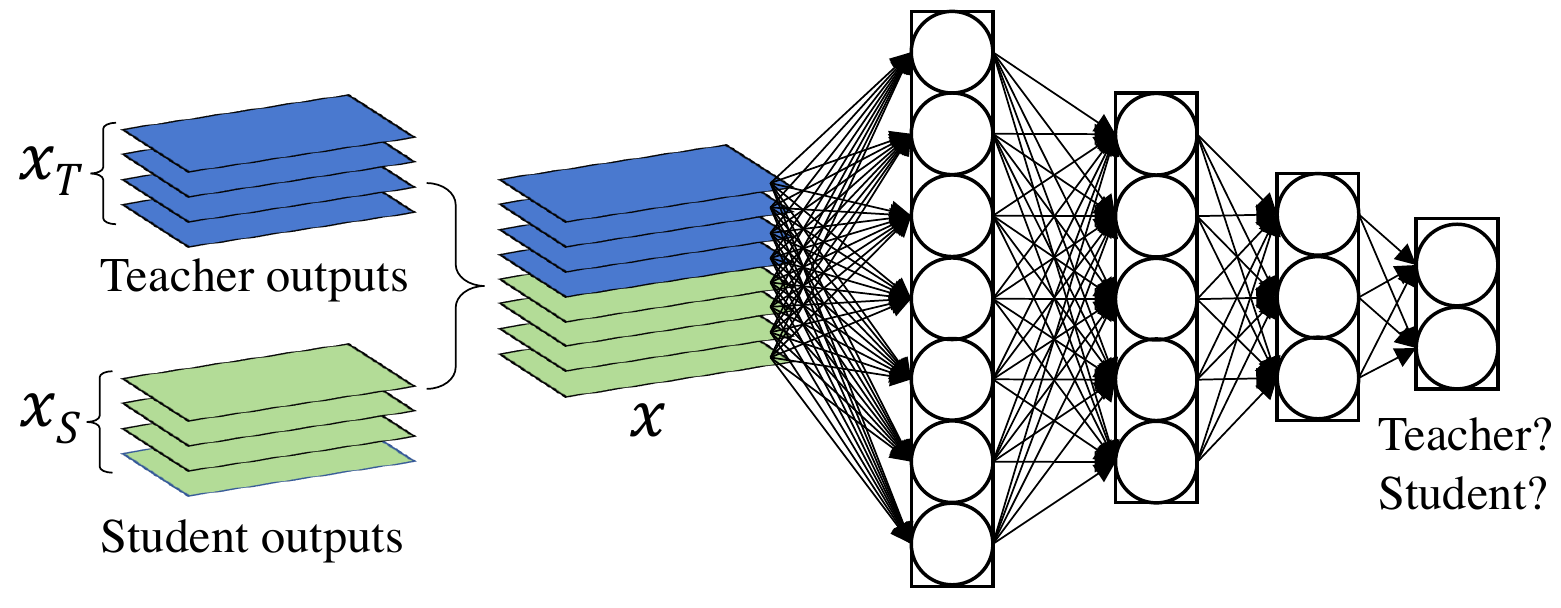}
	\vspace{-0.18in}
	\caption{Illustration of our proposed discriminator. We concatenate the outputs of teacher and student as the inputs of a discriminator. The discriminator is a three-layer fully-connected network.}
	\label{d}
	\vspace{-0.1in}
\end{figure}

\subsection{4.2~~Intermediate Alignment}
\begin{figure}[t]
	\centering
	\includegraphics[width=0.48\textwidth]{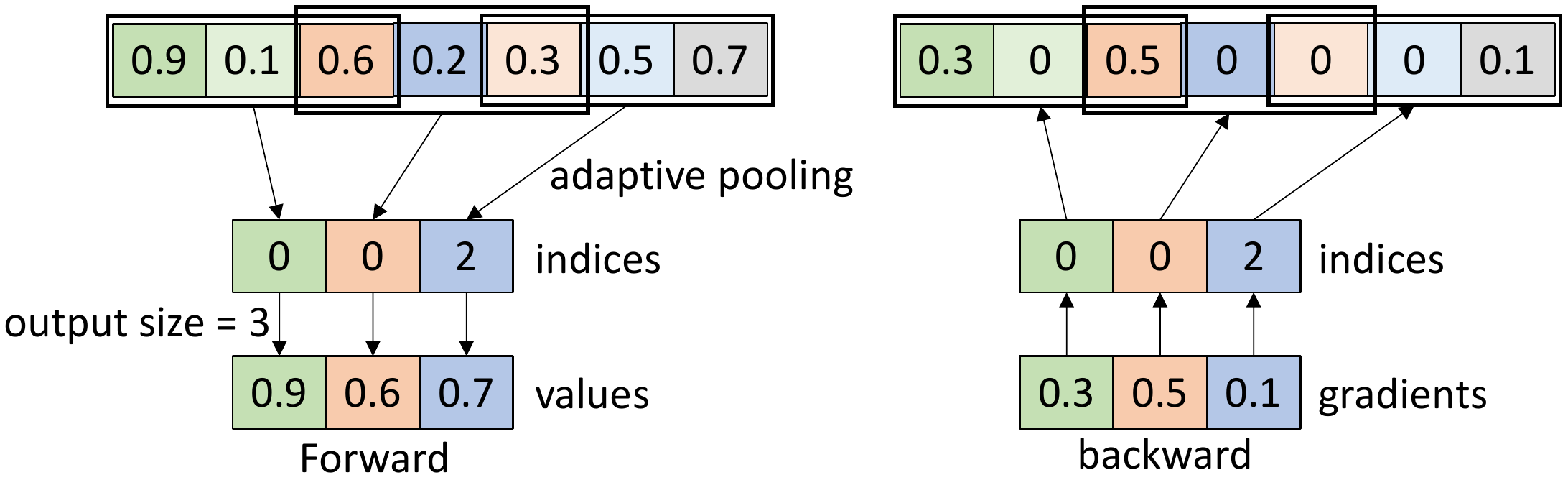}
	\vspace{-0.22in}
	\caption{The process of adaptive pooling in forward and backward stages. We use max operation for illustration.}
	\label{ap}
	\vspace{-0.2in}
\end{figure}

\noindent{\textbf{Adaptive Pooling.}} The purpose of the adaptive pooling layer is to align the intermediate output from teacher network and student network. This kind of layer is similar to the ordinary pooling layer like average or max pooling, but can generate a predefined length of output with different input size. Because of this specialty, we can use the different teacher networks and pool the output to the same length of student output. Pooling layer can also achieve spatial invariance when reducing the resolution of feature maps. Thus, for the intermediate output, our loss function is:
\begin{equation}
\LL^j_{Sim} = d(f(\mathcal{T}_{{\theta}_j}), f(\mathcal{S}_{{\theta}_j}))
\end{equation}

where $T_{{\theta}_j}$ and $S_{{\theta}_j}$ are the outputs at $j$-th layer of the teacher and student, respectively. $f$ is the adaptive pooling function that can be average or max. Fig.~\ref{ap} illustrates the process of  adaptive pooling. Because we adopt multiple intermediate layers, our final similarity loss is a sum of individual one:

\begin{equation} \label{sim}
\LL_{Sim}  = \sum\limits_{j\in {\mathcal{A}} } {\LL^j_{Sim}}
\end{equation}

where $\mathcal{A}$ is the set of layers that we choose to produce output. In our experiments, we use the last layer in each block of a network (block-wise).

\subsection{4.3~~Stacked Discriminators} 

We generate student output by training the student network $\mathcal{S}_\theta$ and freezing the teacher parts adversarially against a series of stacked discriminators {\em D}$_j$. A discriminator {\em D} attempts to classify its input $x$ as teacher or student by maximizing the following objective~\cite{goodfellow2014generative}:
\begin{equation}
\LL^j_{GAN} = \EE_{x\sim p_{\textrm{teacher}}} \log D_j(x) + \EE_{x\sim p_{\textrm{student}}} \log(1 - D_j(x))
\end{equation}

where ${x\sim p_{\textrm{student}}}$ are outputs from generation network $\mathcal{S}_{{\theta}_j}$. At the same time, $\mathcal{S}_{{\theta}_j}$ attempts to generate similar outputs which will fool the discriminator by minimizing $\EE_{x\sim p_{\textrm{student}}} \log(1 - D_j(x))$.


In Eq.~\ref{gan}, $x$ is the concatenation of teacher and student outputs. We feed $x$ into the discriminator which is a three-layer fully-connected network. The whole structure of a discriminator is shown in Fig.~\ref{d}.

\noindent{\textbf{Multi-Stage Discriminators.}}
Using multi-Stage discriminators can refine the student outputs gradually.
As shown in Fig.~\ref{al}, the final adversarial loss is a sum of the individual ones (by minimizing -$\LL^j_{GAN}$):
\begin{equation} \label{gan}
\LL_{GAN}  = -\sum\limits_{j\in {\mathcal{A}}} { \LL^j_{ GAN}}
\end{equation}
Let $\left| \mathcal{A} \right|$ be the number of discriminators. In our experiments, we use 3 for CIFAR~\cite{krizhevsky2009learning} and SVHN~\cite{netzer2011reading}, and 5 for ImageNet~\cite{deng2009imagenet}.

\subsection{4.4~~Joint Training of Similarity and Discriminators}
Based on above definition and analysis, we incorporate the similarity loss in Eq.~\ref{sim} and adversarial loss in Eq.~\ref{gan} into our final loss function. Our whole framework is trained end-to-end by the following objective function: 
\begin{equation}\label{final_loss}
\LL = \alpha \LL_{Sim} + \beta \LL_{GAN}
\end{equation}

where $\alpha$ and $\beta$ are trade-off weights. We set them as 1 in our experiments by cross validation. We also use the weighted coefficients to balance the contributions of different blocks. For 3-block networks, we ues [0.01, 0.05, 1], and [0.001, 0.01, 0.05, 0.1, 1] for 5-block ones.

\section{5.~~Multi-Model Ensemble via Adversarial Learning (MEAL)}
We achieve ensemble with a training method that is simple and straight-forward to implement.
As different network structures can obtain different distributions of outputs, which can be viewed as soft labels (knowledge), we adopt these soft labels to train our student, in order to compress knowledge of different architectures into a single network. Thus we can obtain the seemingly contradictory goal of ensembling multiple neural networks at {\em {no additional testing cost}}. 

\subsection{5.1~~Learning Procedure}
To clearly understand what the student learned in our work, we define two conditions. First, the student has the same structure as the teacher network. Second, we choose one structure for student and randomly select a structure for teacher in each iteration as our ensemble learning procedure.

The learning procedure contains two stages. First, we pre-train the teachers to produce a model zoo. Because we use the classification task to train these models, we can use the softmax cross entropy loss as the main training loss in this stage. Second, we minimize the loss function $\LL$ in Eq.~\ref{final_loss} to make the student output similar to that of the teacher output. The learning procedure is explained below in Algorithm~\ref{inf}.

\begin{algorithm}[h]
	\small
	\caption{Multi-Model Ensemble via Adversarial Learning (MEAL).}
	\label{inf}
	{\bf Stage 1:} 
	
	Building and Pre-training the Teacher Model Zoo $\mathcal{T}=\{\mathcal{T}_\theta^1,\mathcal{T}_\theta^2, \dots \mathcal{T}_\theta^i\}$, including: VGGNet~\cite{simonyan2014very}, ResNet~\cite{he2016deep}, DenseNet~\cite{huang2016densely}, MobileNet~\cite{howard2017mobilenets}, Shake-Shake~\cite{gastaldi2017shake}, etc.
	
	{\bf Stage 2:} 
	\begin{algorithmic}[1]
		\Function{$TSM$}{$\mathcal{T}$}
		\State $\mathcal{T}_\theta \leftarrow RS(\mathcal{T})$ \Comment{Random Selection}
		\State \Return $\mathcal{T}_\theta$
		\EndFunction
		\State	for {\em {each iteration}} do:
		\State \ \ \ \ \ $\mathcal{T}_\theta \leftarrow TSM(\mathcal{T})$
		\Comment{Randomly Select a Teacher Model}
		\State \ \ \ \ \ $\mathcal{S}_\theta = \arg \min _{\mathcal{S}_\theta}\ {\LL}(\mathcal{T}_\theta,\mathcal{S}_\theta)$
		\Comment{Adversarial Learning for a Student}
		\State end for
	\end{algorithmic}
\end{algorithm}

\section{6.~~Experiments and Analysis} \label{exp}
We empirically demonstrate the effectiveness of MEAL on several benchmark datasets. We implement our method on the PyTorch~\cite{paszke2017automatic} platform.
\subsection{6.1.~~Datasets}

{\bf CIFAR.} The two CIFAR datasets~\cite{krizhevsky2009learning} consist of colored natural images with a size of 32$\times$32. CIFAR-10 is drawn from 10 and CIFAR-100 is drawn from 100 classes. In each dataset, the train and test sets contain 50,000 and 10,000 images, respectively. A standard data augmentation scheme\footnote{zero-padded with 4 pixels on both sides, randomly cropped to produce 32x32 images, and horizontally mirror with probability 0.5.}~\cite{lee2015deeply,romero2015fitnets,larsson2016fractalnet,huang2017snapshot,liu2017learning} is used.  We report the test errors in this section with training on the whole training set.

\noindent{\textbf{SVHN.}} The Street View House Number (SVHN) dataset~\cite{netzer2011reading} consists of 32$\times$32 colored digit images, with one class for each digit. The train and test sets contain 604,388 and 26,032 images, respectively. Following previous works~\cite{goodfellow2013maxout,huang2016deep,huang2017snapshot,liu2017learning}, we split a subset of 6,000 images for validation, and train on the remaining images without data augmentation.

\noindent{\textbf{ ImageNet.}} The ILSVRC 2012 classification dataset~\cite{deng2009imagenet} consists of 1000 classes, with a number of 1.2 million training images and 50,000 validation images. We adopt the the data augmentation scheme following~\cite{krizhevsky2012imagenet} and apply the same operation as ~\cite{huang2017snapshot} at test time.


\subsection{6.2~~Networks}
We adopt several popular network architectures as our teacher model zoo, including VGGNet~\cite{simonyan2014very}, ResNet~\cite{he2016deep}, DenseNet~\cite{huang2016densely}, MobileNet~\cite{howard2017mobilenets}, shake-shake~\cite{gastaldi2017shake}, etc. For VGGNet, we use 19-layer with Batch Normalization~\cite{ioffe2015batch}. For ResNet, we use 18-layer network for CIFAR and SVHN and 50-layer for ImagNet. For DenseNet, we use the $BC$ structure with depth L=100, and growth rate k=24. For shake-shake, we use 26-layer 2$\times$96d version. Note that due to the high computing costs, we use shake-shake as a teacher only when the student is shake-shake network.

\begin{table}[h]
	\centering
	\caption{{Ablation study on CIFAR-10 using VGGNet-19 w/BN}. Please
		refer to Section 6.3 for more details.}
			\vspace{1.2ex}
	\resizebox{0.48\textwidth}{!}{%
		\label{ablation}
		\begin{tabular}{c|c|c|c|c|c}
			\hline
			$\ell_1$  \bf dis.    &  $\ell_2$ \bf dis. & \bf {Cross-Entropy}    &   \bf  Intermediate   &   \bf Adversarial  & \bf Test Errors (\%) \\ \hline \hline
			\multicolumn{5}{c|}{\textbf{Base Model (VGG-19 w/ BN)~\cite{simonyan2014very}}} &   {\bf 6.34}    \\ \hline\hline
			\Checkmark	&                                    &                        &                                      &                                     &   6.97  \\ \hline
			&         \Checkmark         &                        &                                      &                                     &    6.22  \\ \hline
			&                                    &    \Checkmark  &                                      &                                     &   6.18   \\ \hline
			&                                    &    \Checkmark  &           \Checkmark         &                                     &   6.10   \\ \hline
			&          \Checkmark        &                        &          \Checkmark          &                                     &   6.17    \\ \hline
			&                                    &    \Checkmark  &          \Checkmark          &        \Checkmark           &  \bf 5.83    \\ \hline 
			\Checkmark  &                                    &                        &          \Checkmark          &        \Checkmark           &    7.57 \\ \hline
		\end{tabular}
	}
\end{table}

\subsection{6.3~~Ablation Studies}
We first investigate each design principle of our MEAL framework. We design several controlled experiments on CIFAR-10 with VGGNet-19 w/BN (both to teacher and student) for this ablation study. A consistent setting is imposed on all the experiments, unless when some components or structures are examined.

\begin{figure*}[t]
	\centering
	\includegraphics[width=0.24\textwidth]{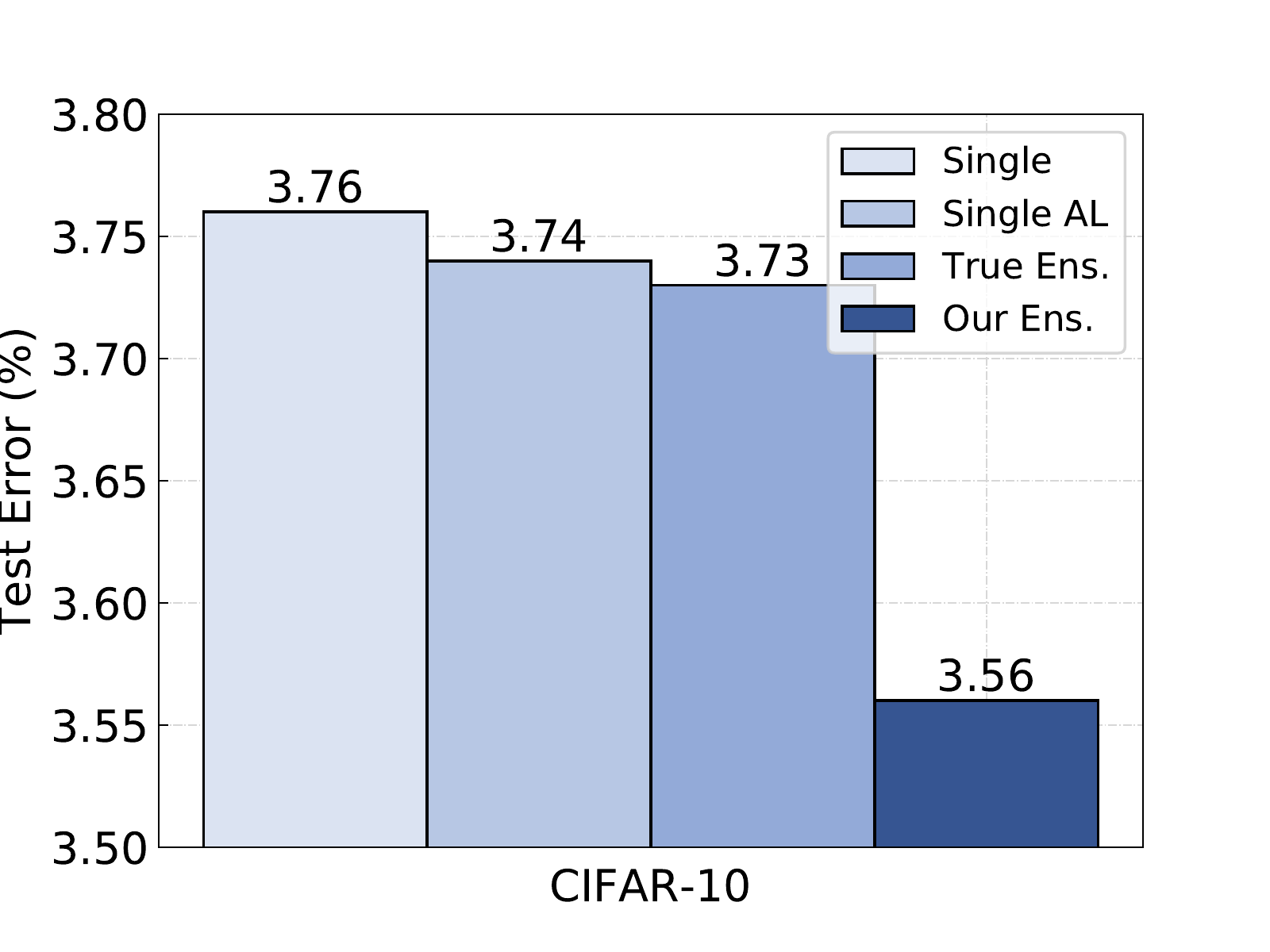}
	\includegraphics[width=0.24\textwidth]{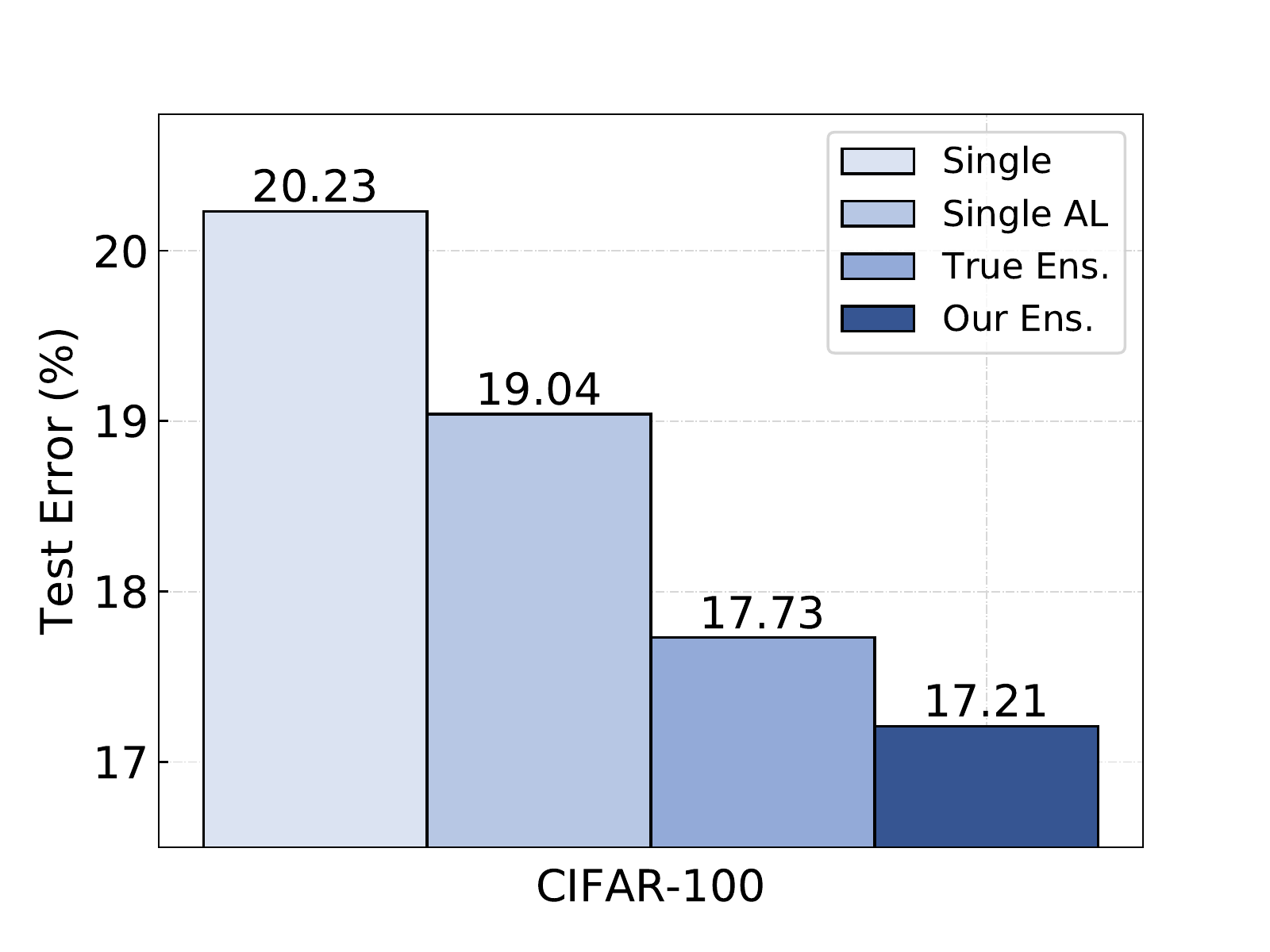}
	\includegraphics[width=0.24\textwidth]{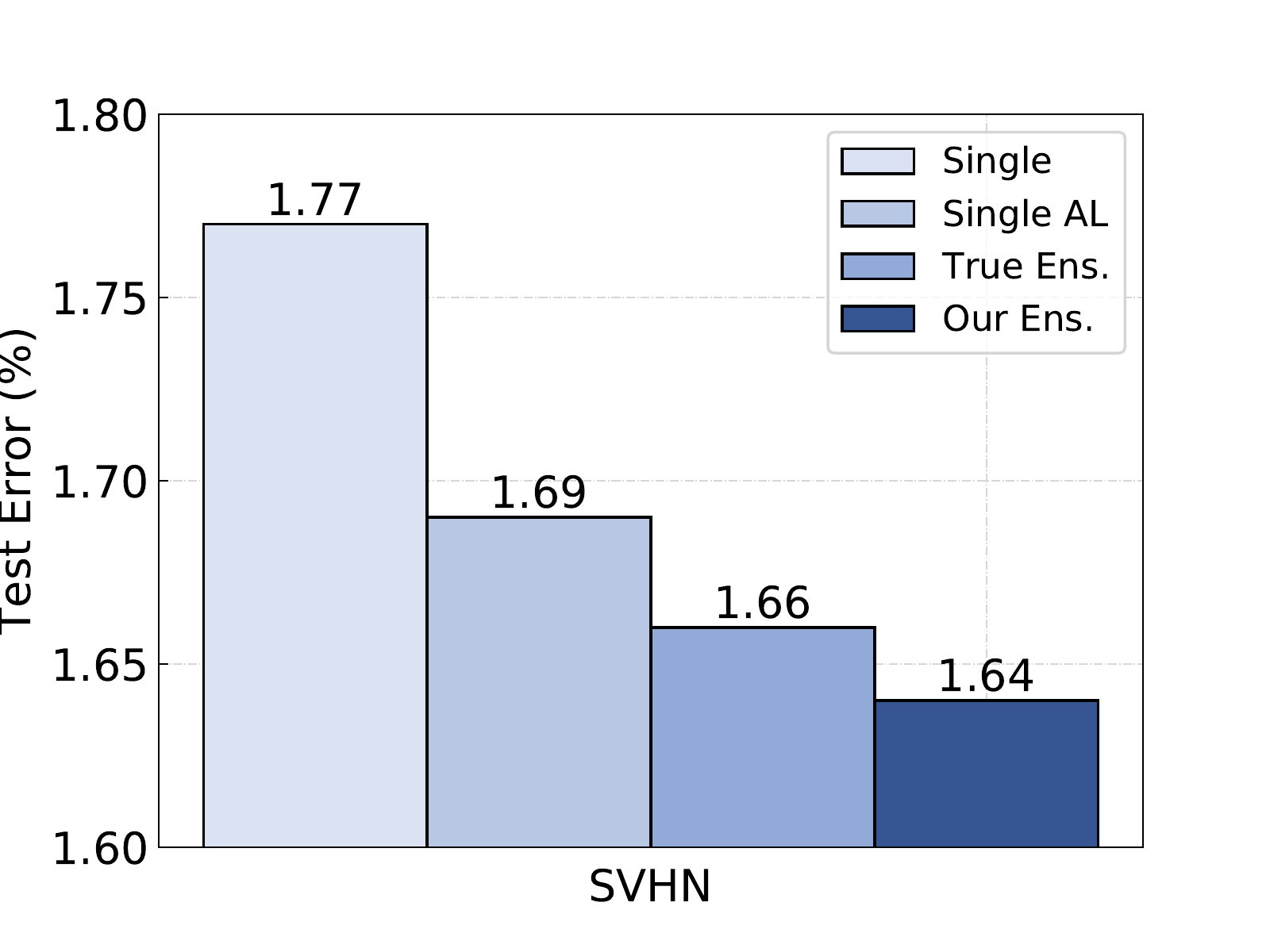}
	\includegraphics[width=0.24\textwidth]{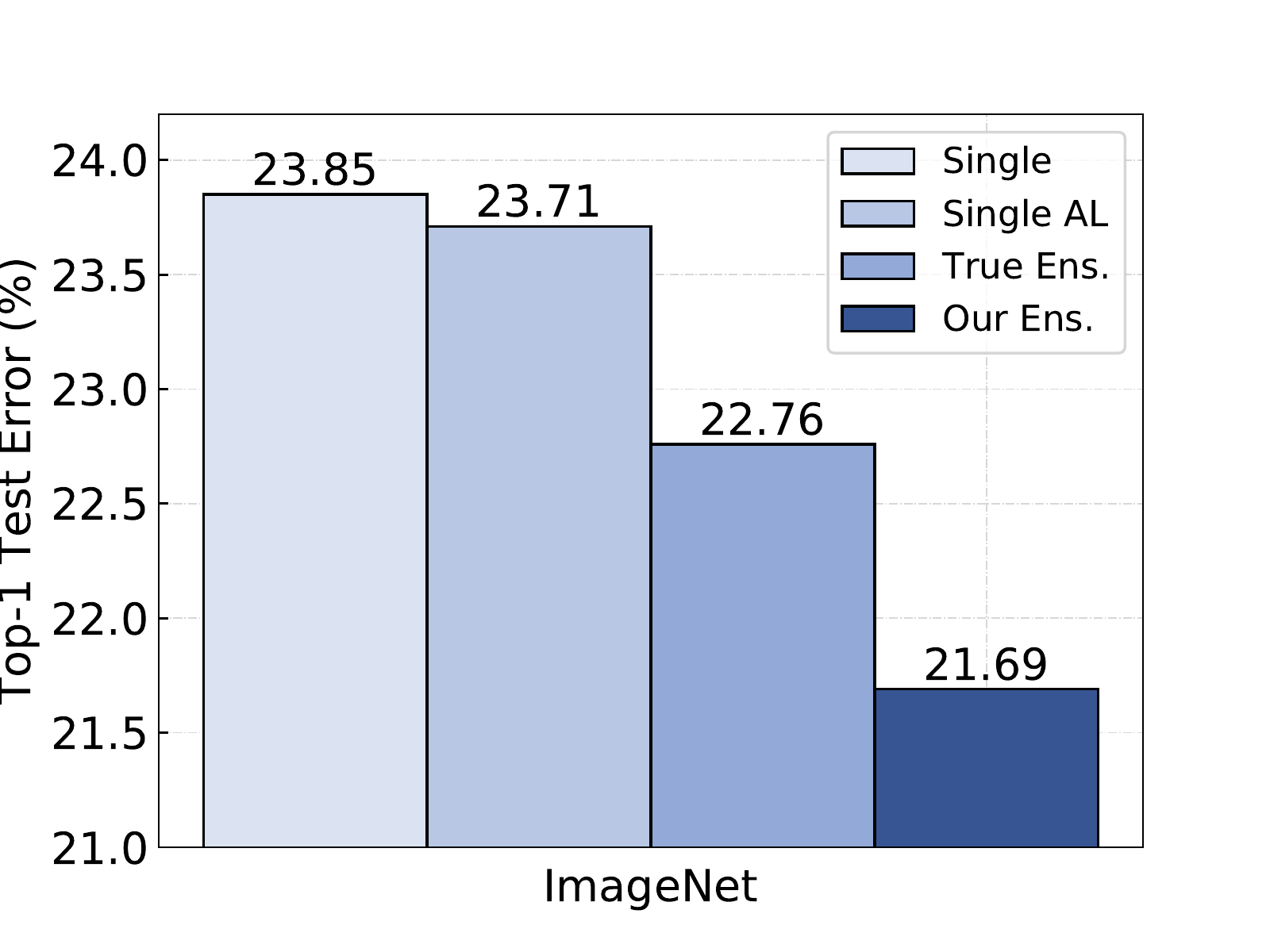}
	\vspace{-0.10in}
	\caption{Error rates (\%) on CIFAR-10 and CIFAR-100, SVHN and ImageNet datasets. In each figure, the results from left to right are 1) base model; 2) base model with adversarial learning; 3) true ensemble/traditional ensemble; and 4) our ensemble results. For the first three datasets, we employ DenseNet as student, and ResNet for the last one (ImageNet).}
	\label{com_all}
\end{figure*}

The results are mainly summarized in Table~\ref{ablation}. The first three rows indicate that we only use $\ell_1$, $\ell_2$ or cross-entropy loss from the last layer of a network. It's similar to the {\em {Knowledge Distillation}} method. We can observe that use cross-entropy achieve the best accuracy. Then we employ more intermediate outputs to calculate the loss, as shown in rows 4 and 5. It's obvious that including more layers improves the performance. Finally, we involve the discriminators to exam the effectiveness of adversarial learning. Using cross-entropy, intermediate layers and adversarial learning achieve the best result. Additionally, we use average based adaptive pooling for alignment. We also tried max operation, the accuracy is much worse (6.32\%).

\subsection{6.4~~Results}
\noindent{\textbf{Comparison with Traditional Ensemble.}} The results are summarized in Figure~\ref{com_all} and Table~\ref{ens}. In Figure~\ref{com_all}, we compare the error rate using the same architecture on a variety of datasets (except ImageNet). It can be observed that our results consistently outperform the single and traditional methods on these datasets. The traditional ensembles are obtained through averaging the final predictions across all teacher models. In Table~\ref{ens}, we compare error rate using different architectures on the same dataset. In most cases, our ensemble method achieves lower error than any of the baselines, including the single model and traditional ensemble.

\begin{table}[h]
	\centering
	\caption{ {Error rate (\%) using different network architectures on CIFAR-10 dataset.}}
	\resizebox{0.48\textwidth}{!}{%
		\label{ens}
		\begin{tabular}{c|c|c|c}
			\toprule
			\bf Network &  \bf Single (\%) & \bf Traditional Ens. (\%) & \bf Our Ens. (\%) \\ \hline\hline
			MobileNet~\cite{howard2017mobilenets}  & 10.70  & -- & \bf 8.09  \\ \midrule 
			VGG-19 w/ BN~\cite{simonyan2014very}  & 6.34  & -- & \bf 5.55  \\ \midrule 
			DenseNet-BC ($k$=24)~\cite{huang2016densely}	&   3.76 &  3.73 & \bf 3.54   \\ \midrule
			Shake-Shake-26 2x96d~\cite{gastaldi2017shake} &  2.86    & 2.79 &  \bf 2.54 \\ \midrule
			\bottomrule
		\end{tabular}
	}
\end{table}	

\noindent{\textbf{Comparison with Dropout.}} We compare MEAL with the ``Implicit'' method Dropout~\cite{srivastava2014dropout}. The results are shown in Table~\ref{drop}, we employ several network architectures in this comparison. All models are trained with the same epochs. We use a probability of 0.2 for drop nodes during training. It can be observed that our method achieves better performance than Dropout on all these networks.

\begin{table}[h]
	\centering
	\caption{ {Comparison of error rate (\%) with Dropout~\cite{srivastava2014dropout} baseline on CIFAR-10.}}
	\resizebox{0.48\textwidth}{!}{%
		\label{drop}
		\begin{tabular}{c|c|c}
			\toprule
			\bf Network &  \bf Dropout (\%)  & \bf Our Ens. (\%) \\ \hline\hline
			VGG-19 w/ BN~\cite{simonyan2014very}  &  6.89 & \bf 5.55  \\ \midrule 
			GoogLeNet~\cite{szegedy2015going} &     5.37  &  \bf 4.83 \\ \midrule
			ResNet-18~\cite{he2016deep} &     4.69  &  \bf 4.35 \\ \midrule
			DenseNet-BC ($k$=24)~\cite{huang2016densely}	&  3.75   & \bf 3.54   \\ \midrule
			\bottomrule
		\end{tabular}
	}
\end{table}	

\noindent{\textbf{Our Learning-Based Ensemble Results on ImageNet.}} As shown in Table~\ref{imagenet}, we compare our ensemble method with the original model and the traditional ensemble. We use VGG-19 w/BN and ResNet-50 as our teachers, and use ResNet-50 as the student. The \#FLOPs and inference time for traditional ensemble are the sum of individual ones. Therefore, our method has both better performance and higher efficiency. Most notably, our MEAL Plus\footnote{denotes using more powerful teachers like ResNet-101/152.} yields an error rate of Top-1 21.79\%, Top-5 5.99\% on ImageNet, far outperforming the original ResNet-50 23.85\%/7.13\% and the traditional ensemble 22.76\%/6.49\%. This shows great potential on large-scale real-size datasets.

\begin{table}[h]
	\centering
	\caption{\bf {Val. error (\%) on ImageNet dataset.}}
	\resizebox{0.48\textwidth}{!}{%
		\label{imagenet}
		\begin{tabular}{c|c|c|c|c}
			\toprule
			Method & Top-1 (\%) & Top-5 (\%) & \#FLOPs & Inference Time (per/image)\\ \hline\hline
			\multicolumn{5}{c}{\textbf{Teacher Networks:}}     \\ \midrule
			VGG-19 w/BN	&   25.76   & 8.15   &  19.52B & $5.70\times10^{-3}$s \\ \midrule
			ResNet-50	&  23.85    & 7.13   &  4.09B & $1.10\times10^{-2}$s\\ \midrule 
			Ours  (ResNet-50)	&    23.58  &  6.86  &  4.09B  &  $ 1.10\times10^{-2}$s \\ \midrule \midrule
			Traditional Ens.  &  22.76    &  6.49  & 23.61B & $1.67\times10^{-2}$s \\ \midrule
			Ours Plus (ResNet-50)	&   \bf 21.79   &\bf 5.99   & \bf 4.09B  &  $\bf 1.10\times10^{-2}$s \\ \bottomrule
		\end{tabular}
	}
\end{table}	

\begin{figure}[h]
	\vspace{-0.1in}
	\centering
	\includegraphics[width=0.35\textwidth]{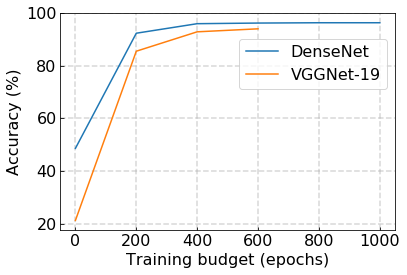}
	\vspace{-0.1in}
	\caption{Accuracy of our ensemble method under different training budgets on CIFAR-10.}
	\label{budget}
	\vspace{-0.1in}
\end{figure}

\begin{figure*}[t]
	\centering
	\includegraphics[width=0.3\textwidth]{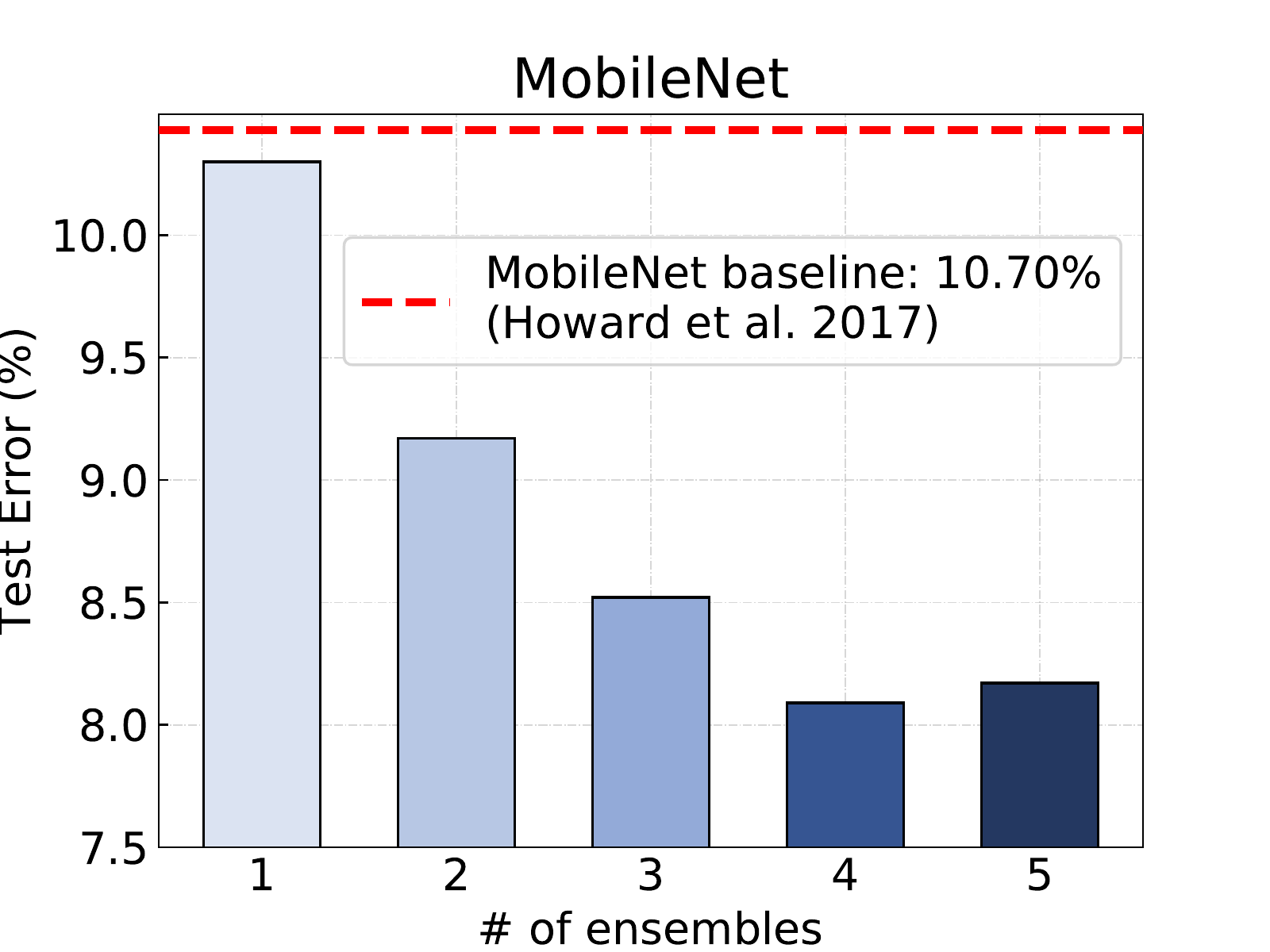}
	\includegraphics[width=0.3\textwidth]{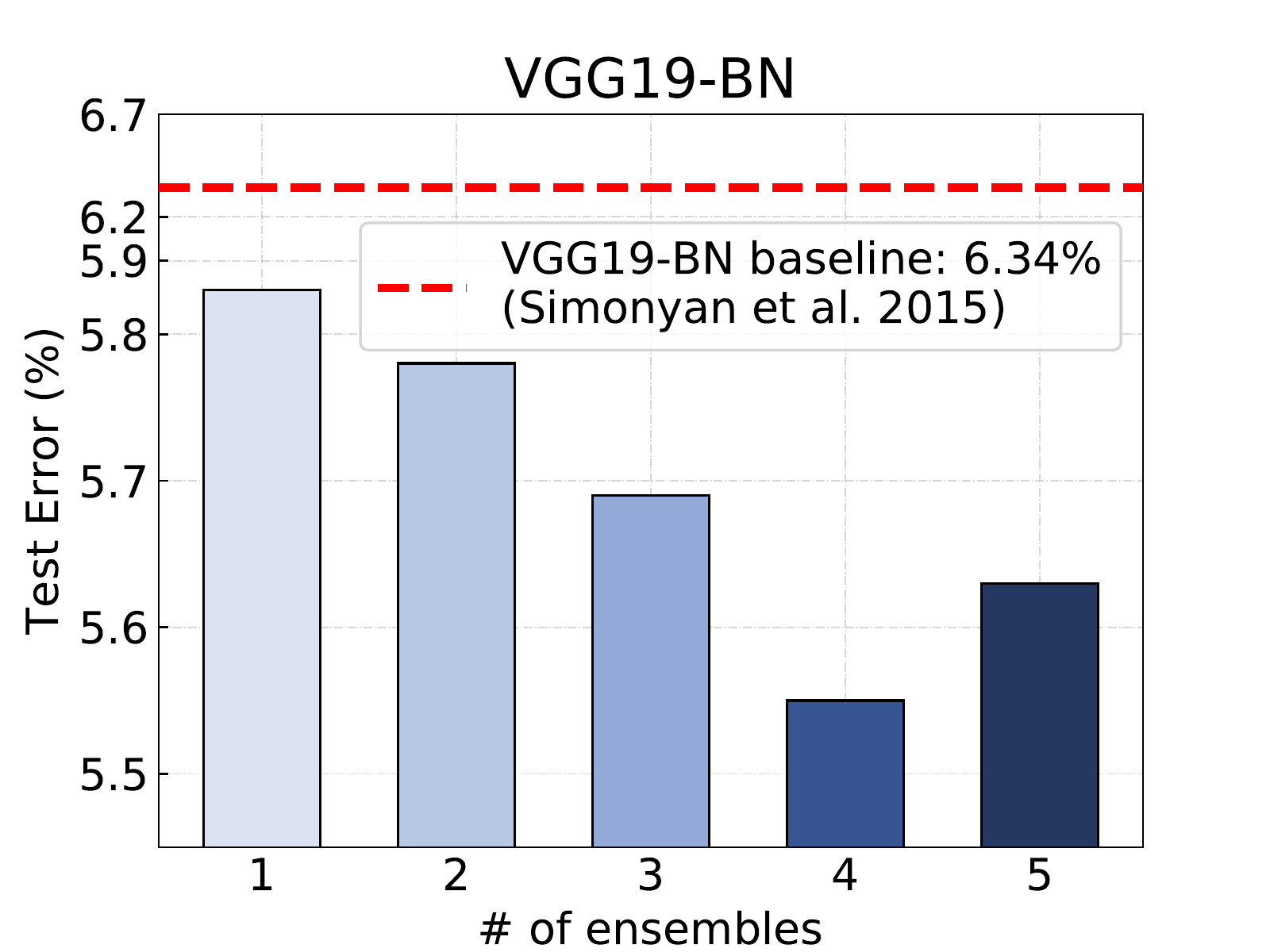}
	\includegraphics[width=0.3\textwidth]{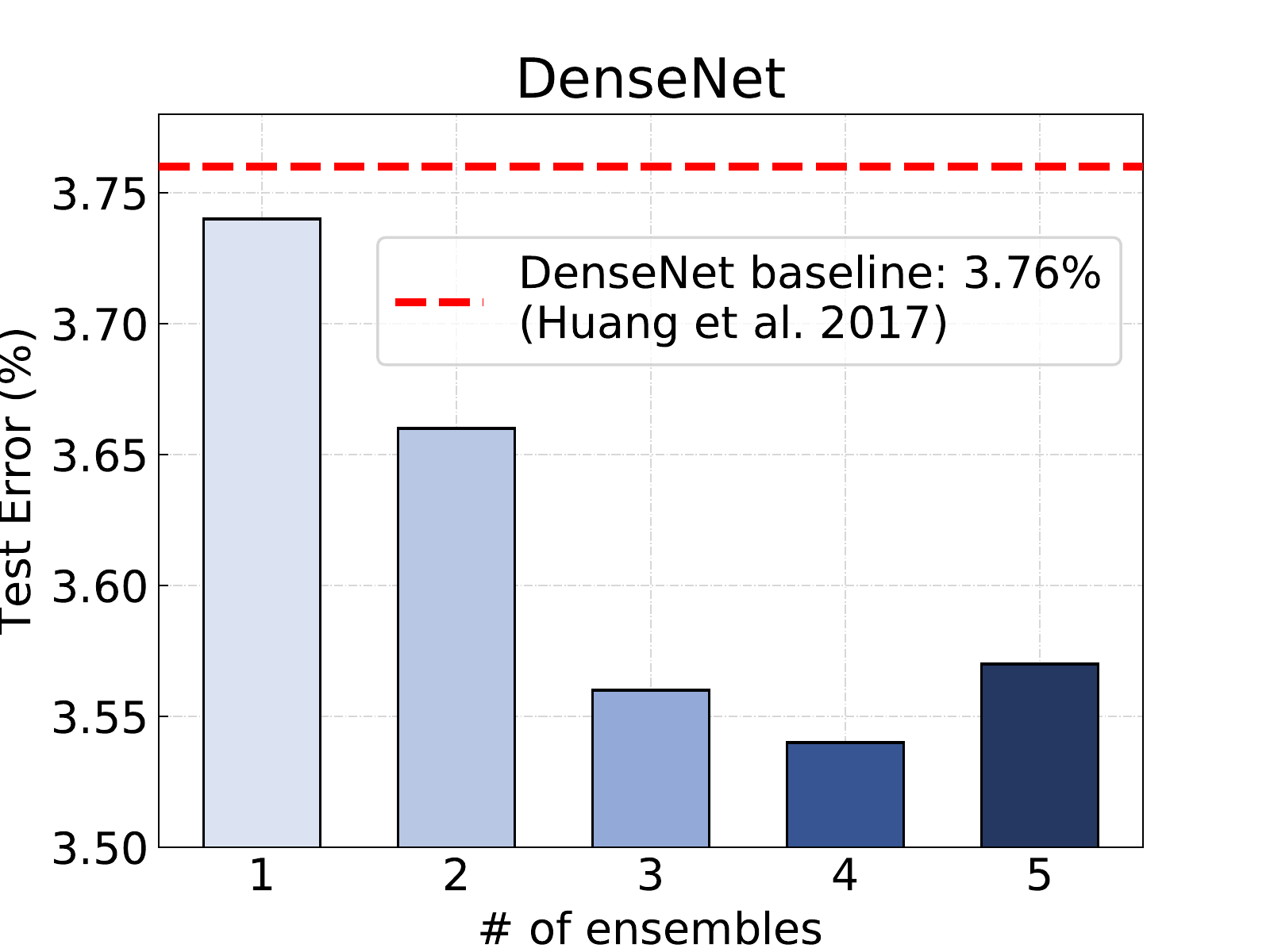}
	\vspace{-0.10in}
	\caption{Error rate (\%) on CIFAR-10 with MobileNet, VGG-19 w/BN and DenseNet.}
	\label{num}
			\vspace{-0.1in}
\end{figure*}

\subsection{6.5~~Analysis}
\noindent{\textbf{Effectiveness of Ensemble Size.}}
Figure~\ref{num} displays the performance of three architectures on CIFAR-10 as the ensemble size is varied. Although ensembling more models generally gives better accuracy, we have two important observations. First, we observe that our single model ``ensemble'' already outputs the baseline model with a remarkable margin, which demonstrates the effectiveness of adversarial learning. Second, we observe some drops in accuracy using the VGGNet and DenseNet networks when including too many ensembles for training. In most case, an ensemble of four models obtains the best performance.

\noindent{\textbf{Budget for Training.}}
On CIFAR datasets, the standard training budget is 300 epochs. Intuitively, our ensemble method can benefit from more training budget, since we use the diverse soft distributions as labels. Figure~\ref{budget} displays the relation between performance and training budget. It appears that more than 400 epochs is the optimal choice and our model will fully converge at about 500 epochs.

\begin{figure}[t]
	\centering
	\includegraphics[width=0.23\textwidth]{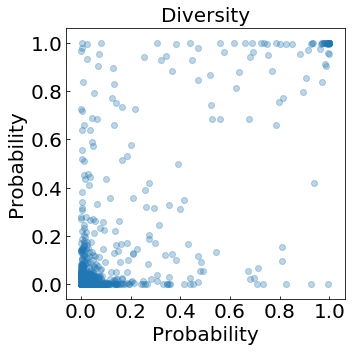}
	\includegraphics[width=0.23\textwidth]{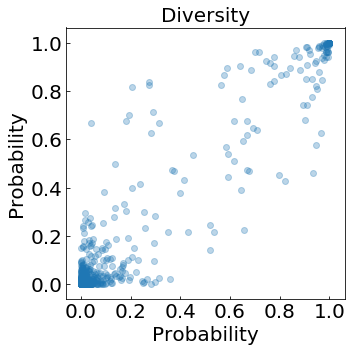}
	\vspace{-0.1in}
	\caption{Probability Distributions between four networks. Left: SequeezeNet {\em{vs.}} VGGNet. Right: ResNet {\em{vs.}} DenseNet.}
	\label{bubble}
	\vspace{-0.1in}
\end{figure}

\noindent{\textbf{Diversity of Supervision.}}
We hypothesize that different architectures create soft labels which are not only informative but also diverse with respect to object categories. We qualitatively measure this diversity by visualizing  the pairwise correlation of softmax outputs from two different networks. To do so, we compute the  softmax predictions for each training image in ImageNet dataset and visualize each pair of the corresponding ones. Figure~\ref{bubble} displays the bubble maps of four architectures. In the left figure, the coordinate of each bubble is a pair of $k$-th predictions ($p^k_{SequeezeNet},p^k_{VGGNet}$), $k=1,2,\dots,1000$, and the right figure is ($p^k_{ResNet},p^k_{DenseNet}$). If the label distributions are identical from two networks, the bubbles will be placed on the master diagonal. 
It's very interesting to observe that the left (weaker network pairs) has bigger diversity than the right (stronger network pairs). It makes sense because the stronger models generally tend to generate predictions close to the ground-truth. In brief, these differences in predictions can be exploited to create effective ensembles and our method is capable of improving the competitive baselines using this kind of diverse supervisions.

\subsection{6.6~~Visualization of the Learned Features}
\begin{figure}[t]
	\centering
	\includegraphics[width=0.2\textwidth]{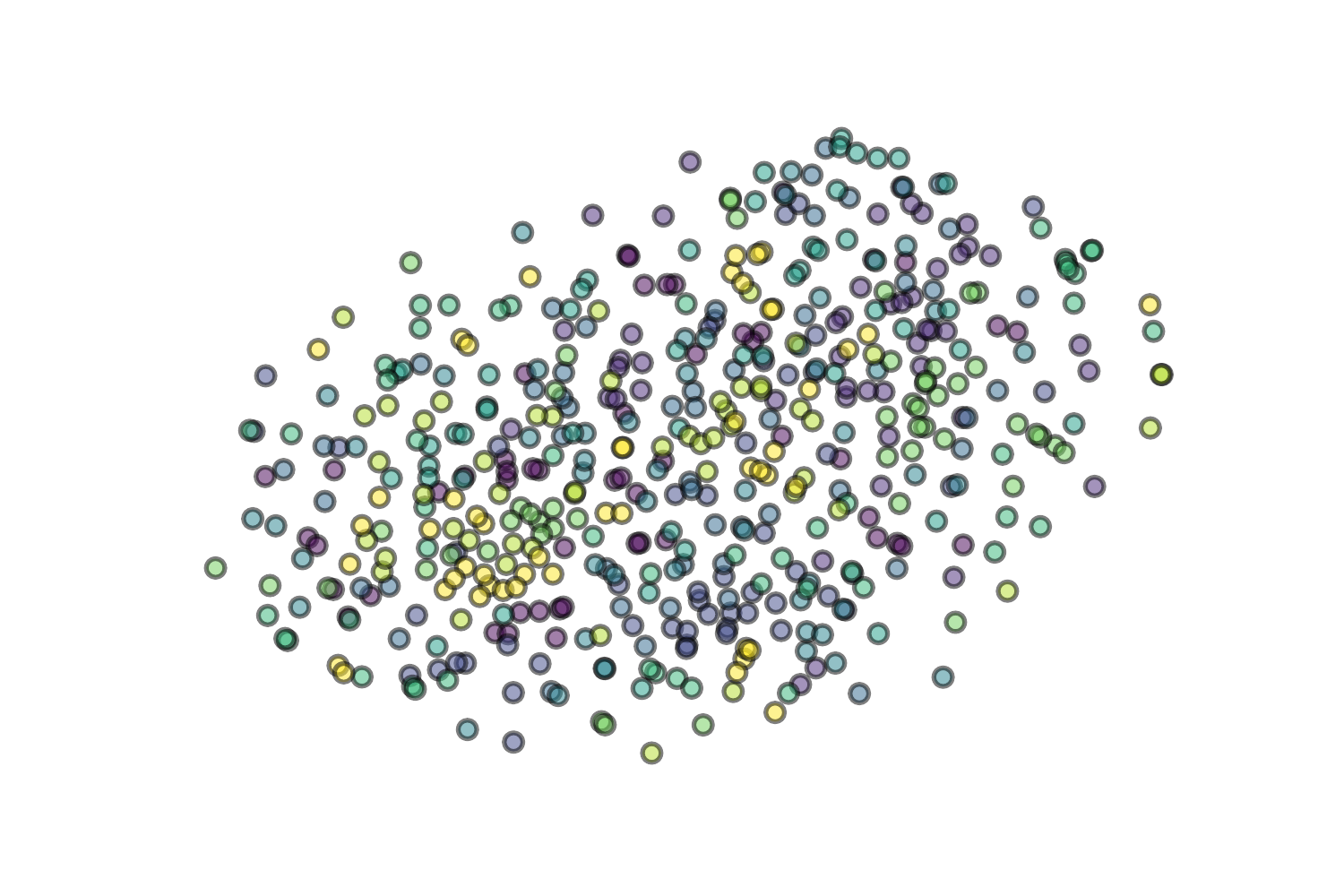}
	\includegraphics[width=0.2\textwidth]{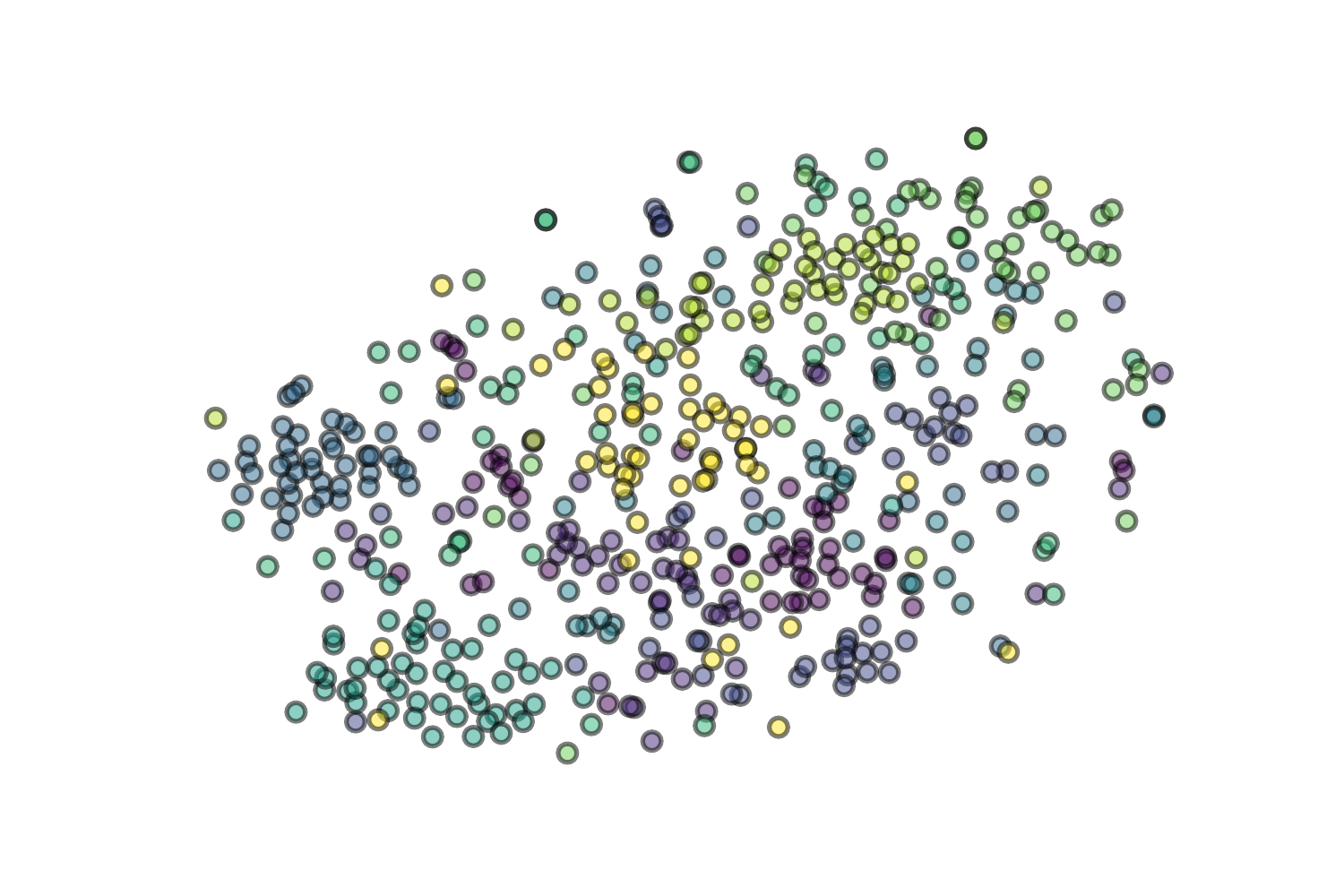}
	\vspace{-0.1in}
	\caption{Visualizations of validation images from the ImageNet dataset by t-SNE~\cite{maaten2008visualizing}. We randomly sample 10 classes within 1000 classes. Left is the single model result using the standard training strategy. Right is our ensemble model result.}
	\label{embedding}
	\vspace{-0.1in}
\end{figure}

To further explore what our model actually learned, we visualize the embedded features from the single model and our ensembling model. The visualization is plotted by t-SNE tool~\cite{maaten2008visualizing} with the last conv-layer features (2048 dimensions) from ResNet-50. We randomly sample 10 classes on ImageNet, results are shown in Figure~\ref{embedding}, it's obvious that our model has better feature embedding result.

\section{7.~~Conclusion}
We have presented MEAL, a learning-based ensemble method that can compress multi-model knowledge into a single network with adversarial learning. Our experimental evaluation on three benchmarks CIFAR-10/100, SVHN and ImageNet verified the effectiveness of our proposed method, which achieved the state-of-the-art accuracy for a variety of network architectures. Our further work will focus on adopting MEAL for cross-domain ensemble and adaption.

\vspace{0.1in}
\noindent{\textbf{Acknowledgements}}
This work was supported in part by National Key R\&D Program of China (No.2017YFC0803700),
NSFC under Grant (No.61572138 \& No.U1611461) and STCSM Project under Grant No.16JC1420400.
{\small
	\bibliographystyle{aaai}
	\bibliography{MEAL_AAAI2019}
}

\end{document}